\documentclass[12pt]{article}

\usepackage{graphicx, url}
\usepackage{amsfonts, amssymb, amscd}
\usepackage{amsmath, delarray, theorem}
\usepackage{float}
\usepackage{natbib}
\bibpunct{(}{)}{;}{a}{}{,}

\newcommand{\beq}{\begin{equation}}
\newcommand{\eeq}{\end{equation}}
\newcommand{\beqn}{\begin{equation*}}
\newcommand{\eeqn}{\end{equation*}}
\newcommand{\beqan}{\begin{eqnarray*}}
\newcommand{\eeqan}{\end{eqnarray*}}

\newcommand{\beqa}{\begin{eqnarray}}
\newcommand{\eeqa}{\end{eqnarray}}
\newcommand{\beu}{\begin{enumerate}}
\newcommand{\eeu}{\end{enumerate}}
\newcommand{\bit}{\begin{itemize}}
\newcommand{\eit}{\end{itemize}}
\newcommand{\bfig}{\begin{figure}}
\newcommand{\efig}{\end{figure}}

\newcommand{\mycap}[1]{\caption{{\small #1}}}
\newcommand{\myvec}[1]{\mathbf{#1}}
\newcommand{\gvec}[1]{\mbox{\boldmath ${#1}$}}

\newcommand{\bvec}{\begin{array}({r})}
\newcommand{\evec}{\end{array}}
\newcommand{\btwomat}{\begin{array}({rr})}
\newcommand{\etwomat}{\end{array}}

\begin{document}

\title{On the Underestimation of Model Uncertainty by Bayesian 
$K$-nearest Neighbors}
\author{Wanhua Su$^1$, Hugh Chipman$^{2,}$\footnote{corresponding author; email: {\tt
hugh dot chipman at acadiau dot ca}},
Mu Zhu$^1$\\[2mm]
$^1$~Department of Statistics and Actuarial Science,\\
University of Waterloo, Waterloo, Ontario, Canada N2L 3G1\\
$^2$~Department of Mathematics and Statistics, \\
Acadia University, Wolfville, Nova Scotia, Canada B4P 2R6}
\maketitle

\begin{abstract}

When using the $K$-nearest neighbors method, one often ignores 
uncertainty in the choice of $K$. To account for 
such uncertainty, \citet{Holmes:Adams:2002} proposed a Bayesian framework 
for $K$-nearest neighbors (KNN). Their Bayesian KNN (BKNN) approach uses a 
pseudo-likelihood function, and standard Markov chain Monte Carlo (MCMC) 
techniques to draw posterior samples. \citet{Holmes:Adams:2002} focused on 
the performance of BKNN in terms of misclassification error but did not 
assess its ability to quantify uncertainty. We present some evidence to 
show that BKNN still significantly underestimates model uncertainty.

\end{abstract}

\section{Introduction} 

The $K$-nearest neighbors method \citep[e.g.,][]{Fix:Hodges:1951, 
Cover:Hart:1967} is conceptually simple but flexible and useful in 
practice. It can be used for both regression and classification. We 
focus on classification only.

\begin{figure} 
\centering 
\includegraphics[width=0.5\textwidth]{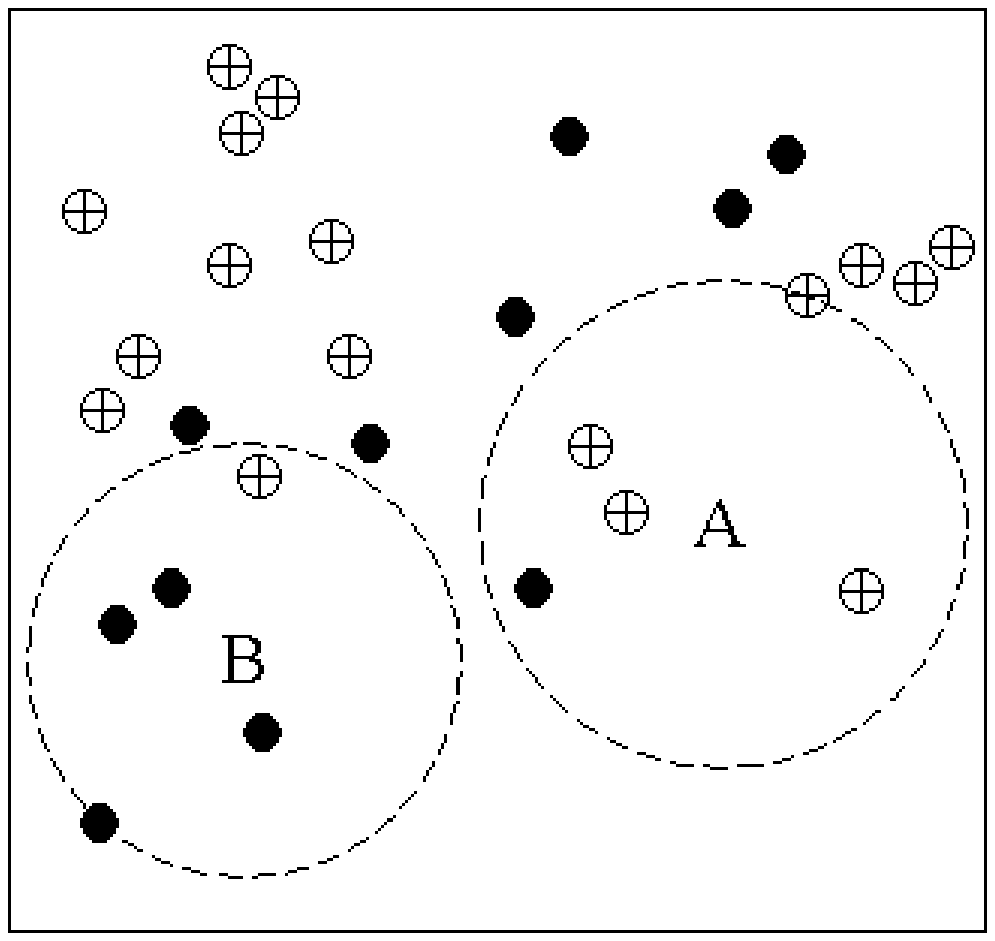}
\mycap{Simulated example illustrating KNN with $K=5$. 
Training observations from class 0 are indicated by the symbol 
``$\oplus$'', and those from class 1 are indicated by the symbol 
``$\bullet$''. A and B are two test points. }
\label{fig:knnplot}
\end{figure}

Under the assumption that points close to one another should have similar 
responses, KNN classifies a new observation according to the class labels 
of its $K$ nearest neighbors. In order to identify the neighbors, one must 
decide how to measure proximity among points and how to define the 
neighborhood. The most commonly-used distance metric is the Euclidean 
distance. The tuning parameter, $K$, is normally chosen by 
cross-validation. Figure \ref{fig:knnplot} 
illustrates how KNN works. Suppose one takes $K=5$. The possible predicted 
values are $\{0/5, 1/5, \cdots, 5/5\}$. Among those five nearest neighbors 
of test point A, four out of five belong to 
class 0. Therefore, A is classified to class 0 with an estimated 
probability of 4/5. Similarly, test point B is 
classified to class 1 with an estimated probability of 4/5.

\citet{Holmes:Adams:2002} pointed out that regular KNN does not account 
for the uncertainty in the choice of $K$. They presented a Bayesian 
framework for KNN (BKNN), compared its performance with the regular KNN on 
several benchmark data sets and concluded that BKNN outperformed KNN in 
terms of misclassification error. By model averaging over the posterior of 
$K$, BKNN is able to improve predictive performance. Unfortunately, they 
never assessed the inferential aspect of BKNN. In this paper, we present some evidence to show that, even 
though BKNN is designed to capture the uncertainty in the choice of $K$, 
it still significantly underestimates overall uncertainty.

\section{BKNN}
\label{sec:bknn}

We first give a quick overview of BKNN in the context of a classification 
problem with $Q$ different classes. To cast KNN into a Bayesian framework, 
\citet{Holmes:Adams:2002} adopted the following (pseudo) likelihood 
function for the data:
\beq 
\label{eq:home}
p(\myvec{Y}|\myvec{X},\beta, K)=
\prod_{i=1}^n p(y_i|\myvec{x}_i, \beta, K)
=\prod_{i=1}^n
\frac{\mbox{exp}\{(\beta/K) \sum_{j\in N(\myvec{x}_i,K)}
\mbox{I}(y_j=y_i)\}}
{\sum_{q=1}^Q\mbox{exp}\{(\beta/K) \sum_{j\in N(\myvec{x}_i,K)}
\mbox{I}(y_j=q)\}}.
\eeq
The indicator function $I$ is 1 whenever its argument is true, and the
notation ``$j\in N(\myvec{x}_i,K)$'' identifies the indices $j$ of the
$K$-nearest neighbours of $\myvec{x}_i$.  Thus 
$\sum_{j\in N(\myvec{x}_i,K)} \mbox{I}(y_j=y_i)$ is $K$ times the 
estimated probability from a conventional KNN model.

There are two unknown parameters, $K$ and $\beta$. The 
parameter $K$ is an positive integer controlling the number of nearest 
neighbors; and $\beta$ is a positive continuous parameter governing the 
strength of interaction between a data point and its neighbors.

The likelihood function (\ref{eq:home}) is a so-called pseudo-likelihood 
function \citep[see, e.g.,][]{Besag:1974, Besag:1975}. Unlike regular 
likelihood functions, the component for data point $y_i$ depends on the 
class labels of other data points $y_j$, for $j \neq i$. Treating $\beta$ 
and $K$ as random variables, the marginal predictive distribution for a 
new data point $(\myvec{x}_{n+1}, y_{n+1})$ based on the training data 
$(\myvec{X}, \myvec{Y})$ is given by
\beq
\label{eq:pred1} 
p(y_{n+1}|\myvec{x}_{n+1},\myvec{X}, \myvec{Y})=
\sum_{K} \int 
p(y_{n+1}|\myvec{x}_{n+1},\myvec{X}, \myvec{Y},\beta,K)
p(\beta, K|\myvec{X}, \myvec{Y}) d \beta, 
\eeq 
where 
\[
p(\beta, K|\myvec{X}, \myvec{Y})\propto p(\myvec{Y}|\myvec{X},
\beta, K)p(\beta, K)
\]
is the posterior distribution of $(\beta, K)$.

Except for the fact that $\beta$ should be positive, little prior 
knowledge is known on the likely values of $K$ and $\beta$. Therefore, 
\citet{Holmes:Adams:2002} adopt independent, non-informative prior 
densities,
\[
p(\beta, K)=p(\beta)p(K)
\]
where
\[
p(K)=\mbox{UNIF}[1, \ldots, K_{\mbox{max}}] \mbox{~~with~~} 
K_{\mbox{max}}=n, \quad
p(\beta)=c\mbox{I}(\beta>0),
\]
and $c$ is a constant so that $p(\beta)$ is an improper flat 
prior on $\mathbb{R}^{+}$.

A random-walk Metropolis-Hastings algorithm is then used to draw $M$ 
samples from the posterior $p(\beta, K|\myvec{X}, \myvec{Y})$, so that 
(\ref{eq:pred1}) can be approximated by
\beqa 
\label{eq:pred2}
p(y_{n+1}|\myvec{x}_{n+1},\myvec{X}, \myvec{Y} ) & \approx &
\frac{1}{M} \sum_{j=1}^M p(y_{n+1}|\myvec{x}_{n+1},\myvec{X},
\myvec{Y}, \beta^{(j)}, K^{(j)}), 
\eeqa 
where $(K^{(j)}, \beta^{(j)})$ is the $j$th sample from the posterior. 

\section{\label{sect.prop}Experiments and results}

One might believe that the Bayesian formulation 
will automatically account for model uncertainty, and that this is a major 
advantage of BKNN over regular KNN. We now describe a simple experiment 
that shows BKNN still significantly underestimates model uncertainty.

The same experiment is repeated 100 times. Each time, we first generate 
$n=250$ pairs of training data from a known, two-class model (details in 
Section~\ref{sect.data}). We then fit BKNN and regular KNN on the training 
data, and let them make predictions at a set of $160$ pre-selected test 
points (details in Section~\ref{sec:testpts}). For each test point, say 
$(\myvec{x}_{n+1}, y_{n+1})$, our parameter of interest is
\beq
\label{eq:keyparam}
 \theta_{n+1} \equiv \Pr(y_{n+1}=1|\myvec{x}_{n+1}).
\eeq
We construct both point estimates (Section~\ref{sec:ptest}) and interval 
estimates (Section~\ref{sec:intest}) of $\theta_{n+1}$:
$\hat\theta_{n+1}$ and $\hat{I}_{n+1}$.

To fit BKNN, we use the Matlab code provided by \citet{Holmes:Adams:2002} 
and exactly the same MCMC setting as described in
\citet[][Section~3.1]{Holmes:Adams:2002}. To fit regular KNN, we use the 
\url{knn} function in R.

\subsection{\label{sect.data}Simulation Model}

\begin{figure}
\centering
\includegraphics[width=0.475\textwidth]{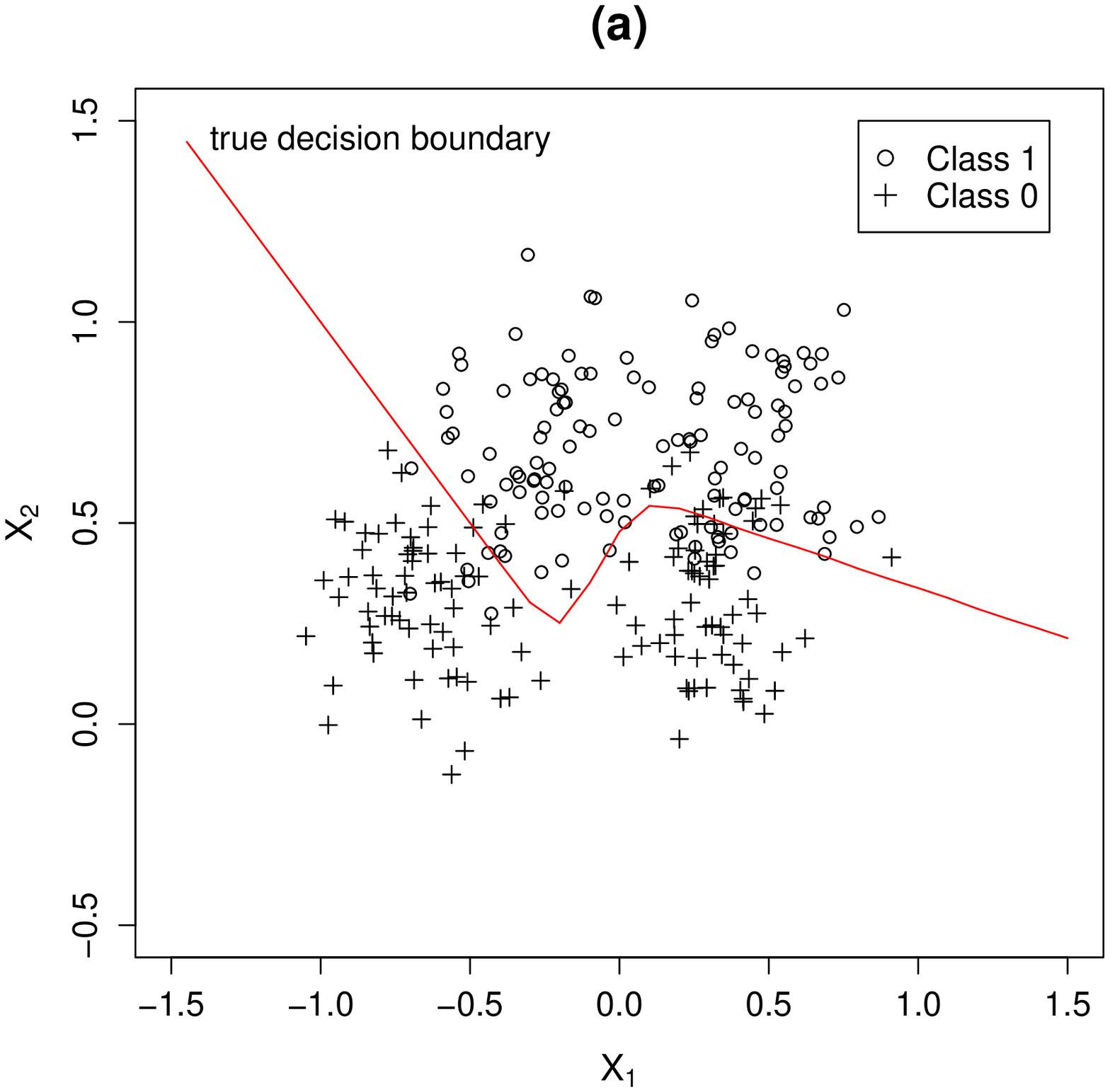}
\includegraphics[width=0.475\textwidth]{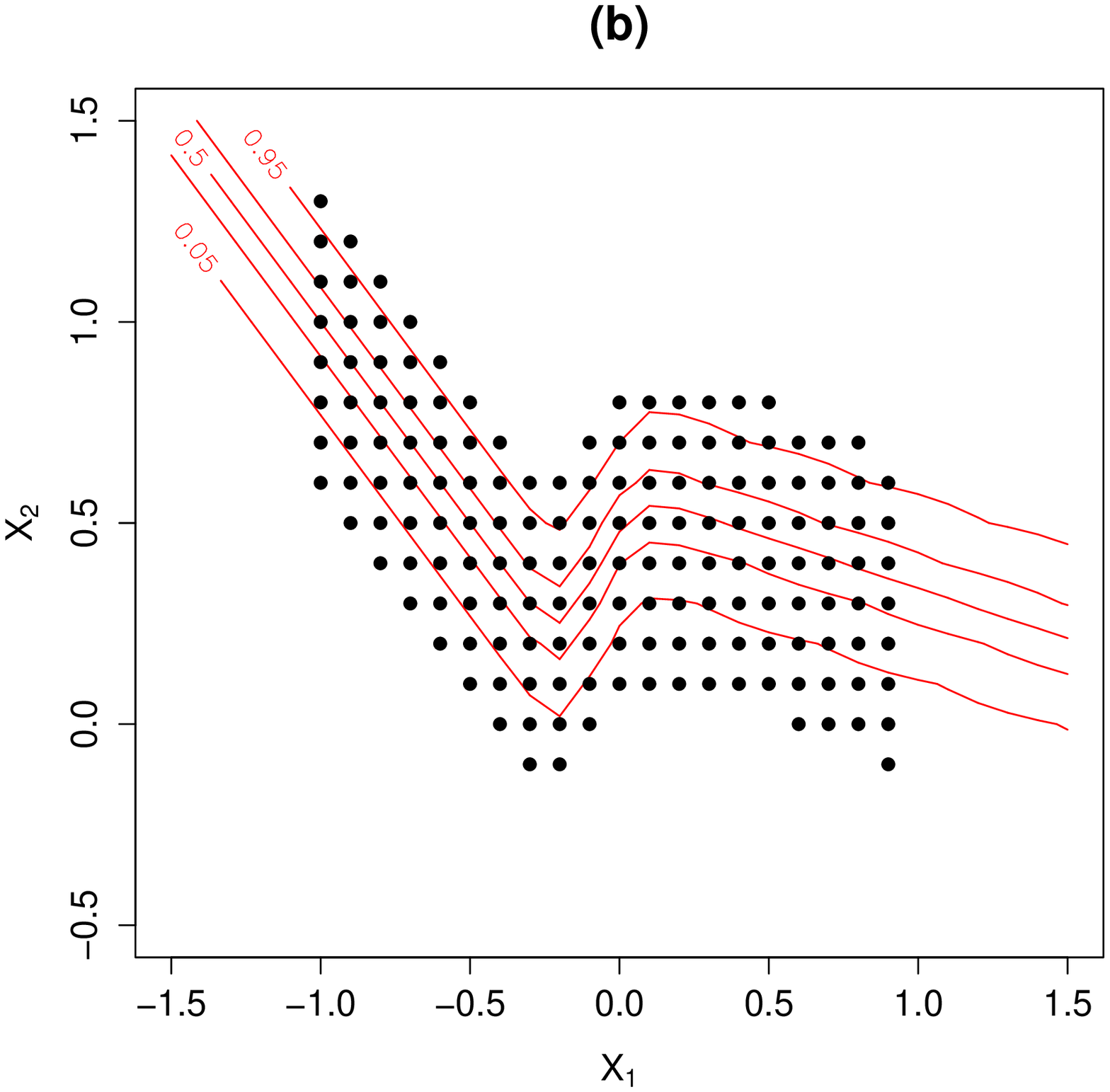}
\mycap{(a) Training data from one experiment, 
and the true probability contour, $\Pr(y=1|\myvec{x})$, as given by 
(\ref{eq:bayes}). (b) The fixed set of test 
points, and the true probability contour. } 
\label{fig:train70} 
\end{figure}

\citet[][Section~3.1]{Holmes:Adams:2002} illustrated BKNN with a synthetic 
dataset consisting of 250 training and 1000 test points, taken from 
\url{http://www.stats.ox.ac.uk/pub/PRNN}. These data were originally 
generated from two classes, each being an equal mixture of two bivariate 
normal distributions. In order to be able to generate slightly different 
training data every time we repeat our experiment, we imitate this 
synthetic data set by assuming the underlying distributions of class 1 
($C_1$) and class 0 ($C_0$) to be:
\beqan
\myvec{x}|C_1 \sim
f_1(\myvec{x}) = 
0.5 \mbox{BVN} \left(\gvec{\mu}_{11}, \gvec{\Sigma} \right)+
0.5 \mbox{BVN} \left(\gvec{\mu}_{12}, \gvec{\Sigma} \right)\phantom{,} \\
\myvec{x}|C_0 \sim
f_0(\myvec{x}) = 
0.5 \mbox{BVN} \left(\gvec{\mu}_{01}, \gvec{\Sigma} \right)+
0.5 \mbox{BVN} \left(\gvec{\mu}_{02}, \gvec{\Sigma} \right),
\eeqan
with 
\[
\gvec{\mu}_{11} =
\bvec 
-0.3  \\ 
\phantom{-}0.7 
\evec, 
\quad
\gvec{\mu}_{12} =
\bvec 
0.4  \\ 
0.7 
\evec, 
\quad
\gvec{\mu}_{01} =
\bvec 
-0.7  \\ 
\phantom{-}0.3 
\evec, 
\quad
\gvec{\mu}_{02} =
\bvec 
0.3 \\ 
0.3
\evec
\]
and
\[
\gvec{\Sigma}=
 \btwomat
 0.03 & 0 \\
 0    & 0.03
 \etwomat.
\] 
The prior class probabilities are taken to be equal, i.e., 
$\Pr(y=1)=\Pr(y=0)=0.5$. Given any data point $\myvec{x}$, its posterior 
probability of being in $C_1$ can be calculated by Bayes' rule
\beq
\label{eq:bayes}
\Pr(y=1|\myvec{x})=\frac{0.5f_1(\myvec{x})}{0.5f_1(\myvec{x})+0.5f_0(\myvec{x})}.
\eeq
Figure~\ref{fig:train70}(a) shows the training data from one experiment 
and the true decision boundary. 

\subsection{Test points}
\label{sec:testpts}

Instead of focusing on the total misclassification error, we focus on 
predictions made at a {\em fixed} set of test points.  These test points 
are chosen as follows: first, we lay out a grid along the first coordinate, 
$X_1 \in \{-1, -0.9, -0.8, \cdots, 0.8, 0.9\}$; for each $X_1$ in that 
grid, eight different values of $X_2$ are chosen so that the test points 
together ``cover'' the critical part of the true posterior probability 
contour. A total of $160$ test points are obtained this way. 
Figure~\ref{fig:train70}(b) shows the fixed set of test points and the 
true posterior probability contour, $\Pr(y=1|\myvec{x})$, as given by 
(\ref{eq:bayes}).

In what follows, we refer to $\theta_{n+1}$ as the key parameter of 
interest, but it should be understood that the subscript ``$n+1$'' is used 
to refer to any test point. There are altogether $160$ such test points, 
and exactly the same calculations are performed for all of them, not just 
one of them.

\subsection{Point estimates of $\theta_{n+1}$}
\label{sec:ptest}

For BKNN, the point estimate of $\theta_{n+1} \equiv 
\Pr(y_{n+1}=1|\myvec{x}_{n+1})$ is the posterior mean:
\[
\hat{\theta}_{n+1}^{BKNN} =
\frac{1}{M} \sum_{j=1}^M \Pr(y_{n+1}=1|\myvec{x}_{n+1}, \myvec{X},
\myvec{Y}, \beta^{(j)}, K^{(j)}), 
\]
where $(K^{(j)}, \beta^{(j)})$ are samples drawn from the posterior 
distribution, $p(K, \beta|\myvec{X}, \myvec{Y})$. For regular KNN, one 
chooses the parameter $K$ by cross-validation, and normally uses the 
original KNN score
\[
 \tilde{\theta}_{n+1}^{KNN} =
 \frac{1}{K} \sum_{j\in N(\myvec{x}_{n+1}, K)}\mbox{I}(y_j=1)
\] 
as the point estimate. In order to make things fully comparable, however, 
we further transform the KNN scores by a logistic model fitted using the 
training data. We describe this next. 

Notice that, for binary classification problems, i.e., $Q=2$, each 
multiplicative term in (\ref{eq:home}) can be rewritten as
\beqa
\label{eq:pknn}
p(y_i|\myvec{x}_i,\beta,K)
&=&\frac{\mbox{exp}\{(\beta/K) \sum_{j\in N(\myvec{x}_i, K)}
\mbox{I}(y_j=y_i)\}}{\mbox{exp}\{(\beta/K) \sum_{j\in N(\myvec{x}_i,K)}
\mbox{I}(y_j=y_i)\}
+\mbox{exp}\{(\beta/K) 
\sum_{j\in N(\myvec{x}_i, K)}
\mbox{I}(y_j \ne y_i)\}}\nonumber\\
&\overset{(\dagger)}{=}&\frac{\mbox{exp}\{(\beta/K) [2\sum_{j\in 
N(\myvec{x}_i,K)}
\mbox{I}(y_j=y_i)-K]\}}
{1+\mbox{exp}\{(\beta/K) [2\sum_{j\in N(\myvec{x}_i, K)}
\mbox{I}(y_j=y_i)-K]\}}\nonumber\\
&=&\frac{\mbox{exp}\{\beta[2g(y_i)-1]\}}{1+\mbox{exp}\{\beta[2g(y_i)-1]\}},
\eeqa 
where 
\beq
\label{eq:KNNscore}
g(y_i)=\frac{1}{K} \sum_{j\in N(\myvec{x}_i, K)}\mbox{I}(y_j=y_i),
\eeq 
is the output of KNN. The step labelled $(\dagger)$ in 
(\ref{eq:pknn}) is due to the identity
\[
\sum_{j\in N(\myvec{x}_i, K)} I(y_j = y_i) + 
\sum_{j\in N(\myvec{x}_i, K)} I(y_j \neq y_i) = K.
\]
Notice that (\ref{eq:pknn}) is equivalent to running a logistic regression 
with no intercept and $[2g(y_i)-1]$ as the only covariate.
Since this extra transformation is built into BKNN, we use 
\beq
\label{eq:KNNptest}
\hat{\theta}_{n+1}^{KNN} = 
\frac{\mbox{exp}\{\hat\beta[2\tilde{\theta}_{n+1}^{KNN}-1]\}}
     {1+\mbox{exp}\{\hat\beta[2\tilde{\theta}_{n+1}^{KNN}-1]\}} 
\eeq
as the point estimate of regular KNN in order to be fully comparable with 
BKNN. In (\ref{eq:KNNptest}), $\hat\beta$ is obtained by running 
a logistic regression of $y_i$ onto $[2g(y_i)-1]$ with no intercept using 
the training data.

\begin{figure}
\centering
\includegraphics[width=0.75\textwidth]{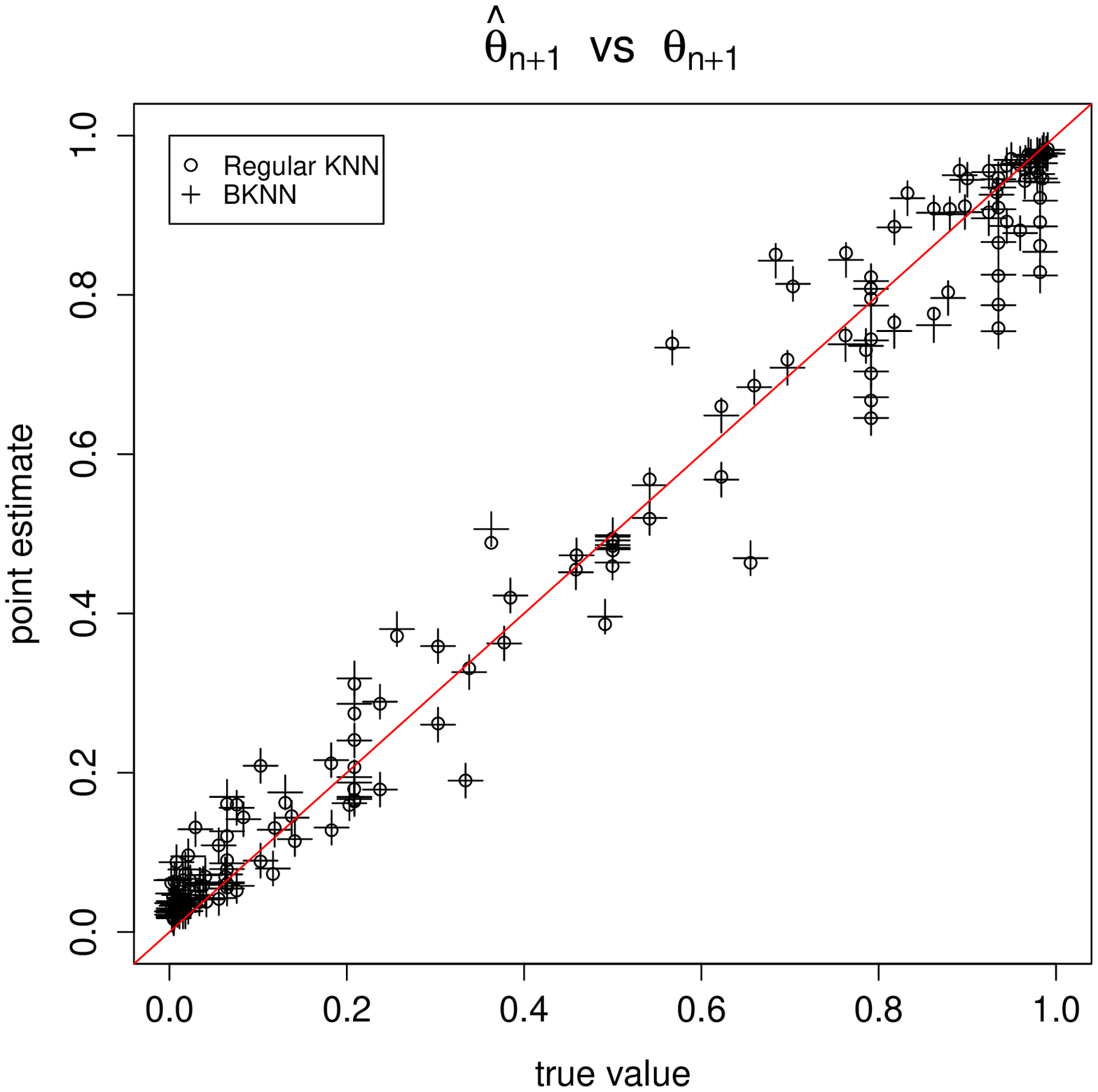}
\mycap{Average of 100 point estimates versus the true parameter value, for 
all $160$ test points. A 45-degree reference line going through the 
origin is also displayed. }
\label{fig:bias} 
\end{figure}

After repeating the experiment 100 times, we obtain 100 slightly different
point estimates at each $\myvec{x}_{n+1}$.
Figure \ref{fig:bias} plots the average of these
100 point estimates against the true value for all $160$ test points. We
see that both BKNN and regular KNN give very similar point estimates.

\subsection{Interval estimates of $\theta_{n+1}$}
\label{sec:intest}

The main focus of our experiments is interval estimation. In 
particular, we are interested in the question of whether these interval 
estimates adequately capture model uncertainty.

For BKNN, we use the 95\% posterior (or credible) interval as our interval 
estimate, $\hat{I}_{n+1}^{BKNN}$. This is constructed by finding the 
$2.5$th and $97.5$th percentiles of the posterior samples. 
To obtain an interval estimate for regular 
KNN, $\hat{I}_{n+1}^{KNN}$, we resort to Efron's bootstrap. Given a 
training set, $\mathcal{D}$, we generate 500 bootstrap samples, 
$\mathcal{D}_1^*, \mathcal{D}_2^*, \cdots, \mathcal{D}_{500}^*$, and 
repeat the entire KNN model building process --- that is, choosing $K$ by 
cross-validation and calculating $\hat\theta_{n+1,b}$ according to 
(\ref{eq:KNNptest}) --- for every $\mathcal{D}_b^*$, 
$b=1,2,\cdots,500$. The interval estimate of $\theta_{n+1}$ is constructed 
by taking the $2.5$th and $97.5$th percentiles of the set, 
$\{\hat\theta_{n+1,1}, \cdots, \hat\theta_{n+1,500}\}$.

\subsubsection{Coverage probabilities}

\begin{figure}
\centering
\includegraphics[width=0.475\textwidth]{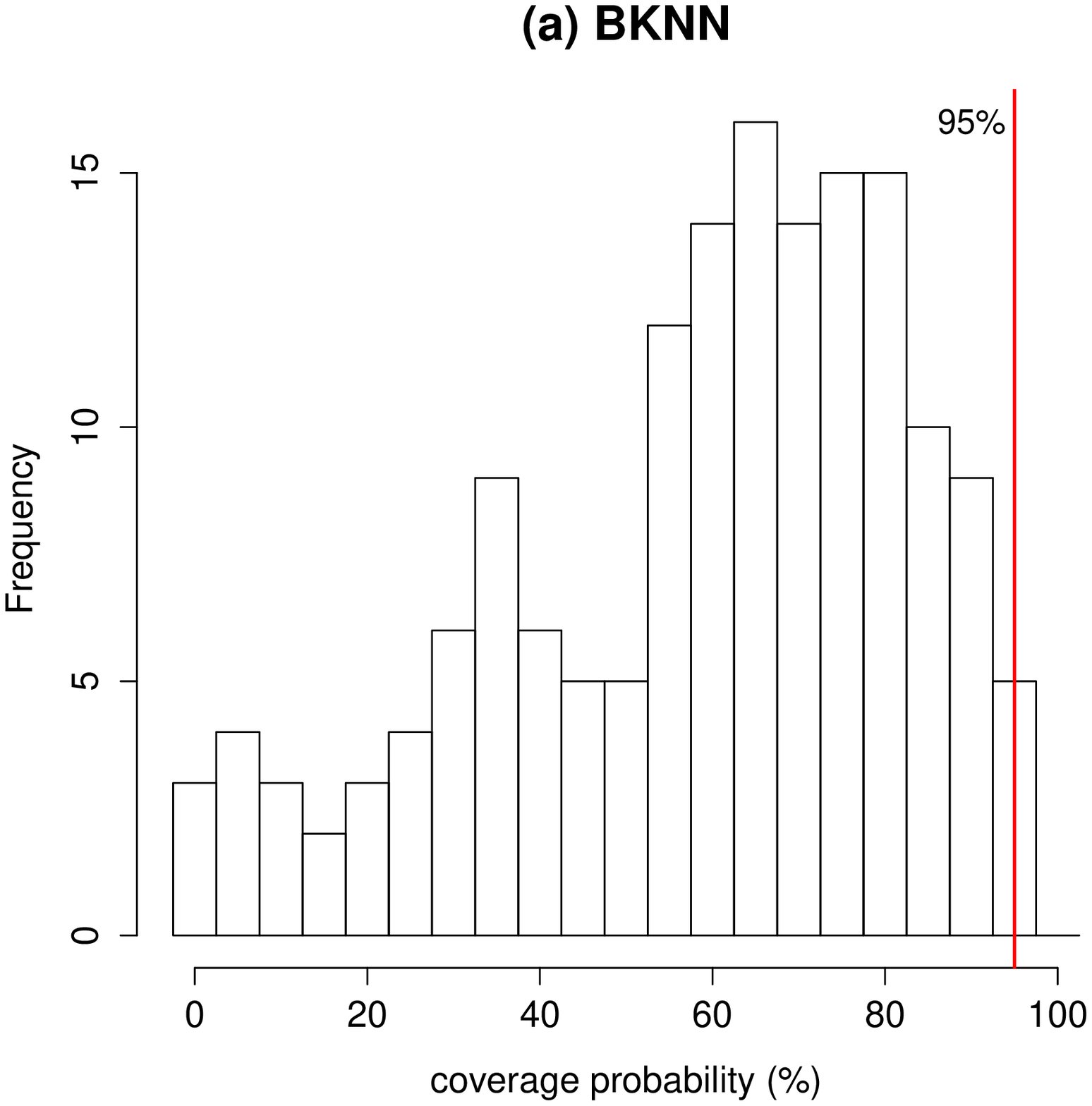}
\includegraphics[width=0.475\textwidth]{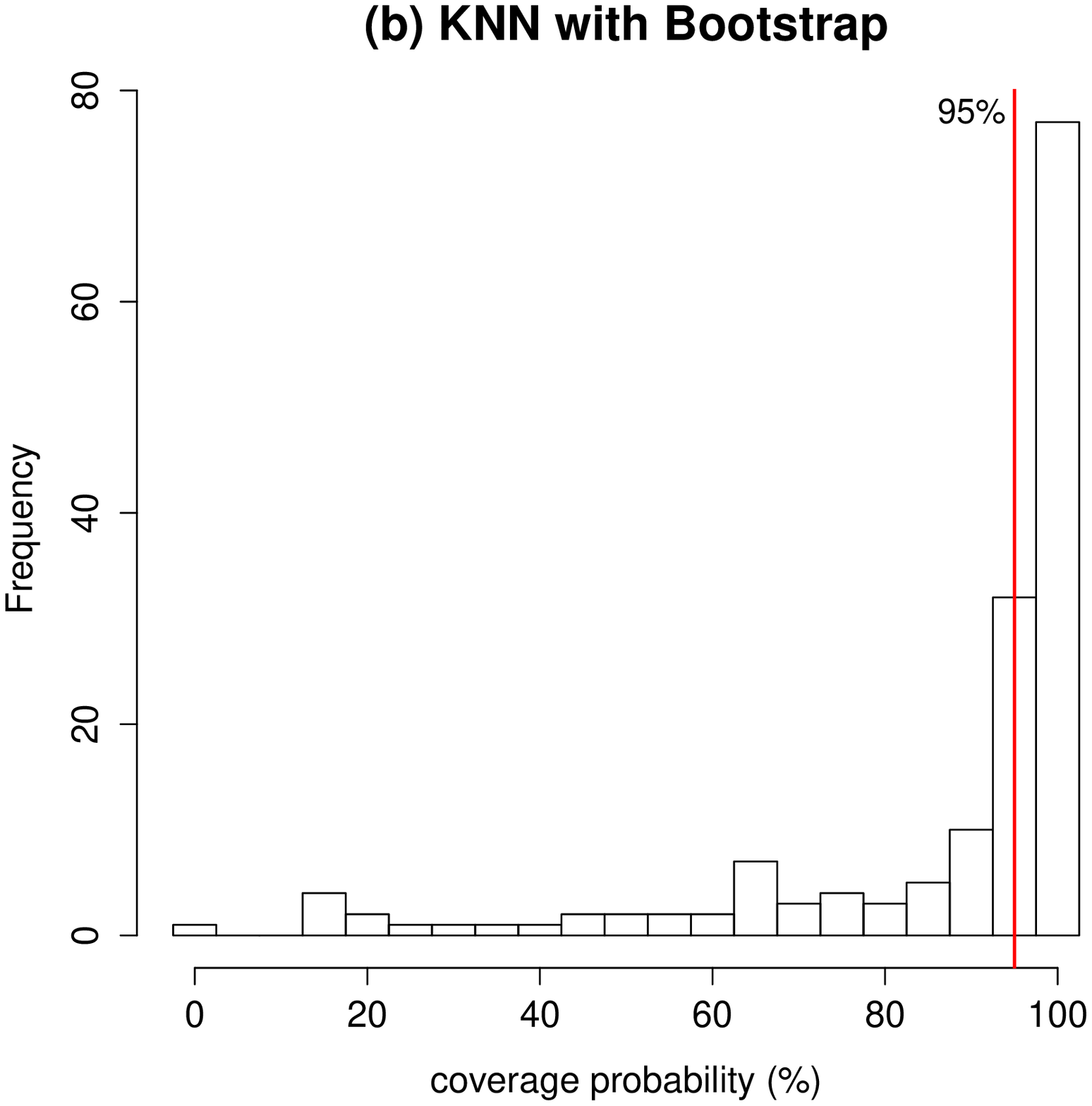}
\mycap{Estimated coverage probabilities of (a) 
$\hat{I}_{n+1}^{BKNN}$ and (b)
$\hat{I}_{n+1}^{KNN}$, for all $160$ test points.} 
\label{fig:coverage} 
\end{figure}

Our first question of interest is: What are the coverage probabilities of 
$\hat{I}_{n+1}^{KNN}$ and $\hat{I}_{n+1}^{BKNN}$?
After repeating the experiment 100 times, we obtain 100 slightly different
interval estimates at each $\myvec{x}_{n+1}$.
The coverage probability of $\hat{I}_{n+1}^{BKNN}$ (and that of 
$\hat{I}_{n+1}^{KNN}$) can be estimated easily by counting the number of 
times $\theta_{n+1}$ is included in the interval over the 100 experiments.
Histograms of the estimated coverage probabilities for all $160$ test 
points are shown in Figure~\ref{fig:coverage}. The posterior intervals 
produced by BKNN can easily be seen to have fairly poor coverage overall.

\subsubsection{Lengths}

For each interval estimate, we can also calculate its length, e.g., 
\beqan
 \mbox{length}^{BKNN}_{n+1} &=& \left| 
 \hat\theta_{n+1}^{BKNN,97.5\%} - 
 \hat\theta_{n+1}^{BKNN,2.5\%} \right|, \\ 
 \mbox{length}^{KNN\phantom{B}}_{n+1} &=& \left| 
 \hat\theta_{n+1}^{KNN,97.5\%\phantom{B}} - 
 \hat\theta_{n+1}^{KNN,2.5\%\phantom{B}} \right|.
\eeqan
Let
\[
\overline{\mbox{length}}^{BKNN}_{n+1}
\quad\mbox{and}\quad
\overline{\mbox{length}}^{KNN}_{n+1}
\]
be the average lengths of these 100 interval estimates. Our second 
question of interest is: Are they too long, too short, or just right? In 
order to answer this question, we need a ``gold standard''.

The very reason for using these interval estimates is to reflect that 
there is uncertainty in our estimate of the underlying parameter, 
$\theta_{n+1}$. This uncertainty is easy to assess directly when one can 
repeatedly generate different sets of training data and repeatedly 
estimate the parameter, which is exactly what we have done. The standard 
deviations of the 100 point estimates (Section~\ref{sec:ptest}), which we 
write as
\[
 \mbox{std}(\hat\theta_{n+1}^{BKNN})
 \quad\mbox{and}\quad
 \mbox{std}(\hat\theta_{n+1}^{KNN}),
\] 
give us a direct assessment of this uncertainty. 

\begin{figure}
\centering
\includegraphics[width=0.75\textwidth]{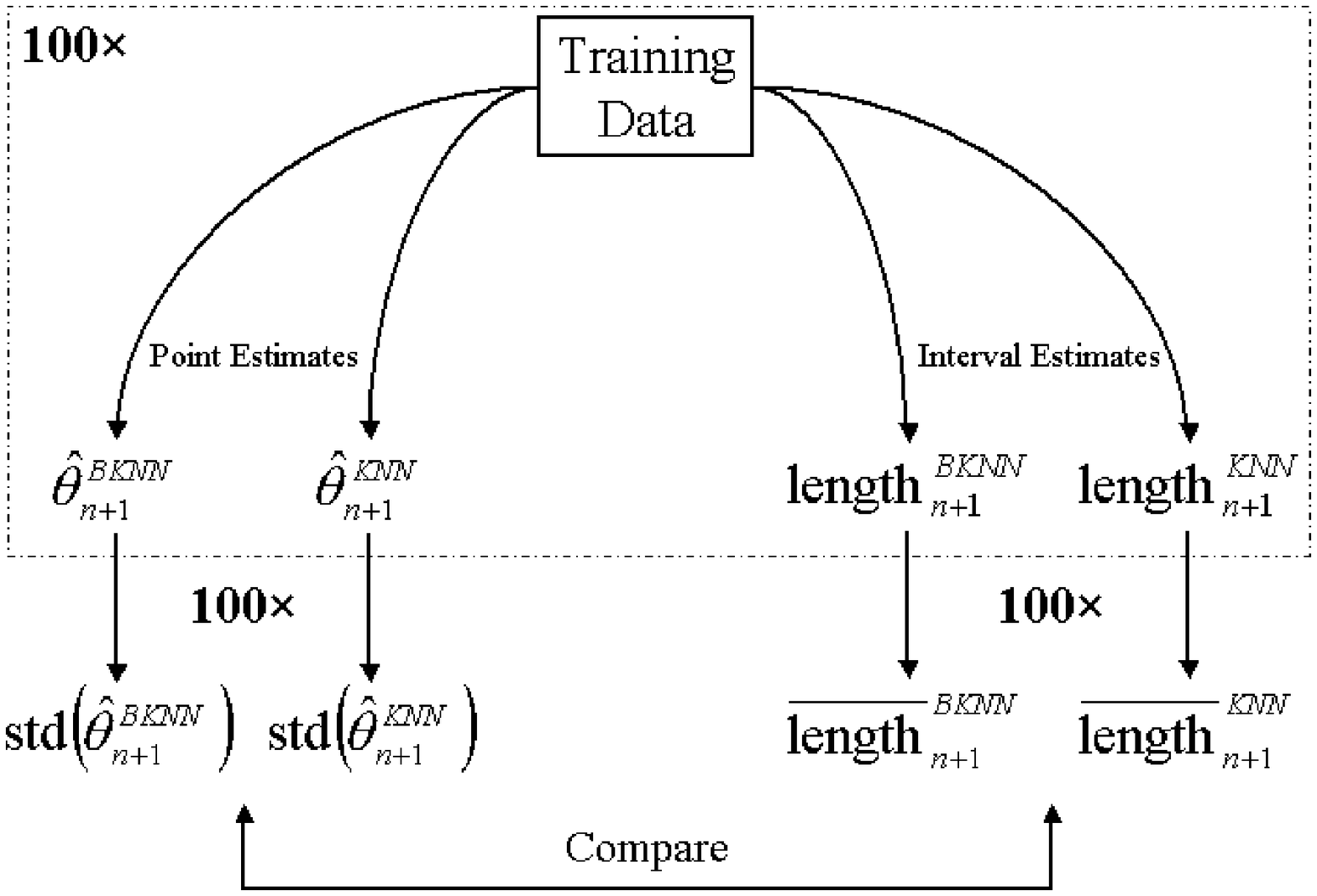}
\mycap{Schematic illustration of our assessment protocol.
Variation over 100 point estimates is used as a benchmark
to assess the quality of the corresponding interval estimates. }
\label{fig:procedure} 
\end{figure}

If the point estimates,
$\hat\theta_{n+1}^{BKNN}$ and
$\hat\theta_{n+1}^{KNN}$, 
are approximately normally distributed, then the correct lengths of the 
corresponding interval estimates should be roughly $4$ times the 
aforementioned standard deviation, that is,
\beqa
\label{eq:goldstd1}
\overline{\mbox{length}}^{BKNN}_{n+1}
&\approx& 4 \times \mbox{std}(\hat\theta_{n+1}^{BKNN}), \\
\label{eq:goldstd2}
\overline{\mbox{length}}^{KNN\phantom{B}}_{n+1}
&\approx& 4 \times \mbox{std}(\hat\theta_{n+1}^{KNN}).
\eeqa
We use (\ref{eq:goldstd1})-(\ref{eq:goldstd2}) as {\em heuristic} 
guidelines to assess how well the interval estimates can capture model 
uncertainty, despite lack of formal justification for the normal 
approximation. 
Figure~\ref{fig:procedure} provides a schematic illustration of our 
assessment protocol.

\begin{figure}
\centering
\includegraphics[width=0.75\textwidth]{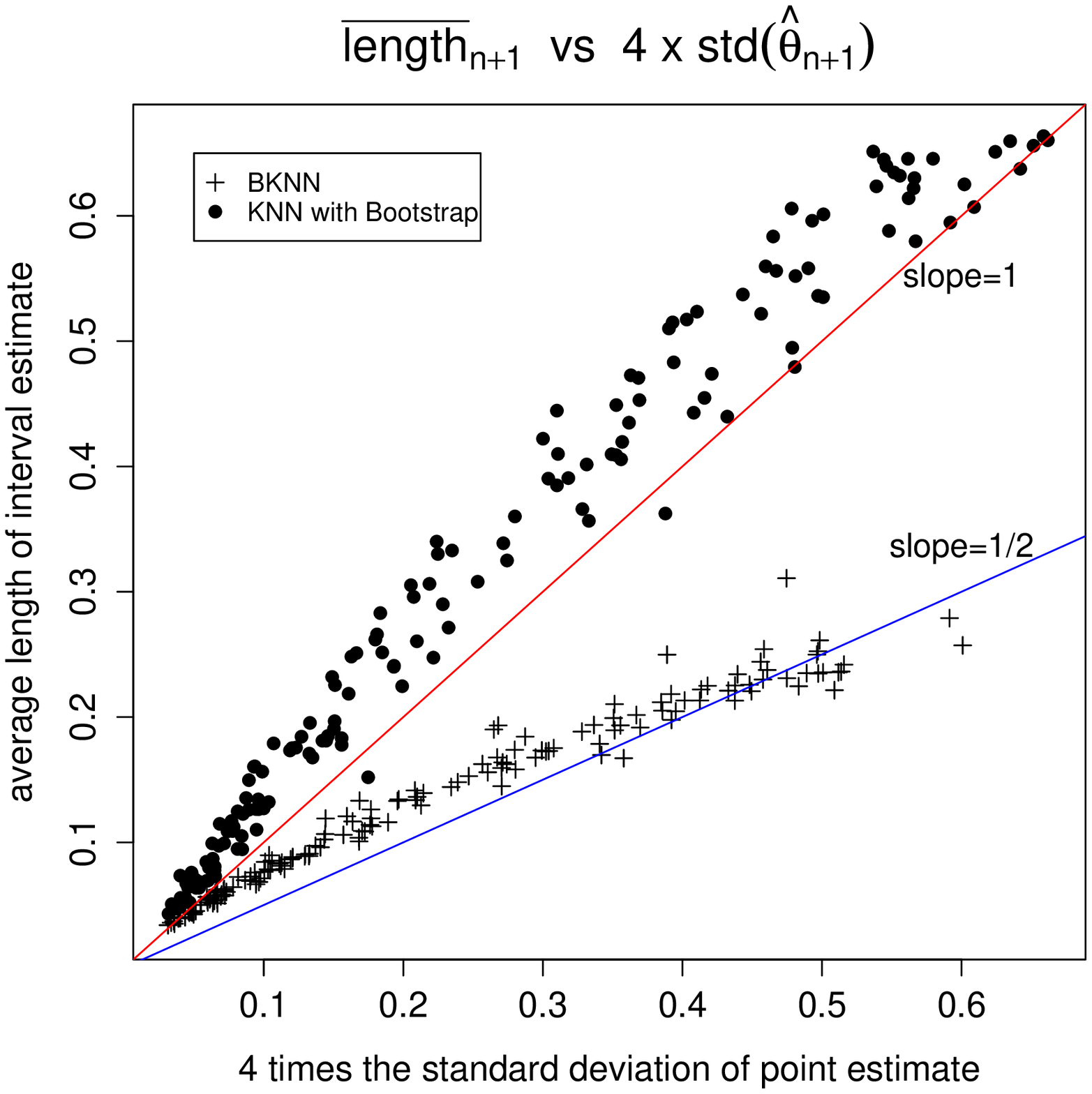}
\mycap{Average length of $100$ interval estimates versus $4$
times the standard deviation of the corresponding point estimate, for all 
$160$ test points. Two reference lines -- both going through the origin, 
one with slope=$1$ and another with slope=$1/2$ --- are 
also displayed.} 
\label{fig:stdvslength} 
\end{figure}

Figure~\ref{fig:stdvslength} plots the average lengths of these 100 
interval estimates against $4$ times the standard deviations of the 
corresponding point estimates --- that is,
$\overline{\mbox{length}}_{n+1}^{BKNN}$ against
$4 \times \mbox{std}(\hat\theta_{n+1}^{BKNN})$ and 
$\overline{\mbox{length}}_{n+1}^{KNN}$ against
$4 \times \mbox{std}(\hat\theta_{n+1}^{KNN})$ 
--- for all $160$ test points. Here, it is easy to see that the Bayesian 
posterior intervals are apparently too short, whereas bootstrapping 
regular KNN gives a more accurate assessment of the amount of uncertainty 
in the point estimate.

\section{\label{sect.discussion}Discussion}

Why does BKNN underestimate uncertainty? We believe it is because BKNN 
only accounts for the uncertainty in the {\em number} of neighbors (i.e., 
the parameter $K$), but it is unable to account for the uncertainty in the 
spatial {\em locations} of these neighbors. This is a general phenomenon 
associated with pseudo-likelihood functions.

Pseudo-likelihood functions were first introduced by \citet{Besag:1974, 
Besag:1975} to model spatial interactions in lattice systems. Since then, 
they have been widely used in image processing \citep[e.g.,][]{Besag:1986} 
and network tomography \citep[e.g.,][]{Strauss:Ikeda:1990, 
Liang:Yu:2003, Robins:2007}. However, statistical inference based on 
pseudo-likelihood functions is still in its infancy. Some researchers 
argue that pseudo-likelihood inference can be problematic since it ignores 
at least part of the dependence structure in the data. In applications to 
model social networks, a number of researchers, such as 
\citet{Wasserman:Robins:2005} and \citet{Snijders:2002}, have pointed out 
that maximum pseudo-likelihood estimates are substantially biased and the 
standard errors of the parameters are generally underestimated. For BKNN, 
the pseudo-likelihood function (\ref{eq:home}) clearly ignores the fact 
that the locations of one's neighbors are also random, not just the number 
of neighbors.

However, for complex networks whose full likelihood functions are 
intractable, models based on pseudo-likelihood are attractive (if not the 
only) alternatives \citep{Strauss:Ikeda:1990}. Rather than trying to write 
down the full likelihood functions for these difficult problems, it is 
probably more fruitful to concentrate our research efforts on how to 
adjust or correct standard error estimates produced by the 
pseudo-likelihood. To this effect, one interesting observation from 
Figure~\ref{fig:stdvslength} is the fact that 
\[ 
\overline{\mbox{length}}_{n+1}^{BKNN} \approx 
2 \times \mbox{std}(\hat\theta_{n+1}^{BKNN}).
\] 
If we continue to use $4 \times \mbox{std}(\hat\theta_{n+1}^{BKNN})$ as 
the ``gold standard'', then these Bayesian posterior intervals are about 
half as long as they should be. We have observed this phenomenon on other 
examples, too, but do not yet have an explanation for it.

Despite the fact that BKNN seems to underestimate overall uncertainty, 
that $\overline{\mbox{length}}_{n+1}^{BKNN}$ is still approximately 
proportional to $\mbox{std}(\hat\theta_{n+1}^{BKNN})$ suggests that we can 
still rely on it to assess the {\em relative} uncertainty of its 
predictions. For many practical problems, this is still very useful. For 
example, if two accounts, A and B, are both predicted to be fraudulent 
with a high probability of $0.9$ but the posterior interval of A is twice 
as long as that of B, then it is natural for a financial institution to 
spend its limited resources investigating account B rather than account A.

\section*{Acknowledgment}

This research is partially supported by the Natural Science and 
Engineering Research Council (NSERC) of Canada, Canada's National 
Institute for Complex Data Structures (NICDS) and the Mathematics of 
Information Technology And Complex Systems (MITACS) network. 

\bibliographystyle{/u/m3zhu/natbib} 
\bibliography{myrefer}

\begin{thebibliography}{}

\bibitem[Besag(1974)]{Besag:1974}
Besag, J. (1974).
\newblock Spatial interaction and the statistical analysis of lattice systems
  (with discussion).
\newblock {\em Journal of Royal Statistical Society: Series B}, {\bf 36}(2),
  192--236.

\bibitem[Besag(1975)]{Besag:1975}
Besag, J. (1975).
\newblock Statistical analysis of non-lattice data.
\newblock {\em The Statistician}, {\bf 24}(3), 179--195.

\bibitem[Besag(1986)]{Besag:1986}
Besag, J. (1986).
\newblock On the statistical analysis of dirty pictures.
\newblock {\em Journal of Royal Statistical Society: Series B}, {\bf 48}(3),
  259--302.

\bibitem[Cover and Hart(1967)]{Cover:Hart:1967}
Cover, T. and Hart, P. (1967).
\newblock Nearest neighbor pattern classification.
\newblock {\em IEEE Transactions on Information Theory}, {\bf IT-13}, 21--27.

\bibitem[Fix and Hodges(1951)]{Fix:Hodges:1951}
Fix, E. and Hodges, J.~L. (1951).
\newblock Discriminatory analysis--nonparametric discrimination: Consistency
  properties.
\newblock Technical report, USAF School of Aviation Medicine, Randolph Field,
  Texas.

\bibitem[Holmes and Adams(2002)]{Holmes:Adams:2002}
Holmes, C.~C. and Adams, N.~M. (2002).
\newblock A probabilistic nearest neighbour method for statistical pattern
  recognition.
\newblock {\em Journal of Royal Statistical Society: Series B}, {\bf 64}(2),
  295--306.

\bibitem[Liang and Yu(2003)]{Liang:Yu:2003}
Liang, G. and Yu, B. (2003).
\newblock Maximum pseudo likelihood estimation in network tomography.
\newblock {\em IEEE Transactions on Signal Processing}, {\bf 51}, 2043--2053.

\bibitem[Robins {\em et~al.}(2007)]{Robins:2007}
Robins, G., Pattison, P., Kalish, Y., and Lusher, D. (2007).
\newblock An introduction to exponential random graph ({$P^{\ast}$}) models for
  social networks.
\newblock {\em Social Networks}, {\bf 29}, 173--191.

\bibitem[Snijders(2002)]{Snijders:2002}
Snijders, T. A.~B. (2002).
\newblock Markov chain monte carlo estimation of exponential random graph
  model.
\newblock {\em Journal of Social Structure}, {\bf 3}(2).

\bibitem[Strauss and Ikeda(1990)]{Strauss:Ikeda:1990}
Strauss, D. and Ikeda, M. (1990).
\newblock Pseudolikelihood estimation for social networks.
\newblock {\em Journal of the American Statistical Association}, {\bf 85},
  204--212.

\bibitem[Wasserman and Robins(2005)]{Wasserman:Robins:2005}
Wasserman, S.~S. and Robins, G.~L. (2005).
\newblock An introduction to random graphs, dependence graphs, and
  {$p^{\ast}$}.
\newblock In J.~S.~P. Carrrington and S.~S. Wasserman, editors, {\em Models and
  Methods in Social Network Analysis}, pages 148--161. Cambridge University
  Press, Cambridge.

\end{thebibliography}

\end{document}